\documentclass{article}

\usepackage{spconf}
\usepackage{graphicx}
\usepackage{amssymb}
\usepackage{epstopdf}
\usepackage{algorithmic}
\usepackage{subfig}
\usepackage{pseudocode}
\usepackage{fancybox}
\usepackage{multicol}
\usepackage{verbatim}
\usepackage{amsthm}
\usepackage{amsmath}
\usepackage{color}
\usepackage{bm}
\usepackage{cite}
\usepackage{multirow}
\usepackage{flushend}
\usepackage{tikz}

\newcommand{\mbf}[1]{{\mathbf #1}}

\newcommand{\bs}{\boldsymbol}

\newtheorem{theorem}{Theorem}[section]

\newtheorem{proposition}[theorem]{Proposition}

\title{Super-resolution MRI using finite rate of innovation curves}
\name{Greg Ongie$^{\star}$ \qquad Mathews Jacob$^{\dagger}$\thanks{This work is supported by grants NSF CCF-0844812, NSF CCF-1116067,  NIH 1R21HL109710-01A1, ACS RSG-11-267-01-CCE, and ONR-N000141310202.}}%
\address{$^{\star}$ Department of Mathematics, University of Iowa, IA, USA. \\
		$^{\dagger}$ Department of Electrical and Computer Engineering, University of Iowa, IA, USA.}

\begin{document}
%
\maketitle
%

\begin{abstract}
We propose a two-stage algorithm for the super-resolution of MR images from their low-frequency k-space samples. In the first stage we estimate a resolution-independent mask whose zeros represent the edges of the image. This builds off recent work extending the theory of sampling signals of finite rate of innovation (FRI) to two-dimensional curves. We enable its application to MRI by proposing extensions of the signal models allowed by FRI theory, and by developing a more robust and efficient means to determine the edge mask. In the second stage of the scheme, we recover the super-resolved MR image using the discretized edge mask as an image prior. We evaluate our scheme on simulated single-coil MR data obtained from analytical phantoms, and compare against total variation reconstructions. Our experiments show improved performance in both noiseless and noisy settings.
\end{abstract}

\section{INTRODUCTION}

The availability of high-resolution MRI data can greatly facilitate early diagnosis by enabling the detection and characterization of subtle and clinically significant lesions \cite{van2012super}. However, the recovery of very high-resolution MRI data is often challenging, mainly due to the slow nature of MRI acquisition, subject motion, and the rapid decrease in SNR with resolution. For example, it is common practice to acquire low-resolution data in MR spectroscopic imaging since the acquisition of higher k-space samples comes with a heavy SNR penalty, which is undesirable for imaging metabolites at very low concentrations. At the same time, the use of low spatial resolution results in the leakage of strong signals such as water and fat to other spatial regions, thus distorting the metabolite signals. 

Several super-resolution and off-the-grid methods were introduced recently to estimate parametric signals with finite number of unknowns from their low-frequency samples. For example, the locations and amplitudes of a finite number of Dirac delta functions can be reliably estimated from its low-frequency Fourier samples using continuous total variation minimization \cite{candes2014towards}, or by estimating an annihilation filter and then determining its roots \cite{vetterli2002sampling}. Unfortunately, the representation using finite number of basis functions is not efficient for the representation of piecewise polynomial images. For example, the gradient of a piecewise constant image is non-zero on a curve, which cannot be represented as a finite linear combination of Dirac delta functions. 

Recently, Pan et al.\ introduced an complex analytic signal model for continuous domain two-dimensional signals, whose derivatives are supported on a curve \cite{pan2013sampling}. Specifically, they assume the curve is the zero level-set of a function band-limited to a rectangular region in the Fourier domain. The authors showed that the Fourier transform of the curve model will annihilate the Fourier coefficients of the signal derivatives. This property enables them to extend the FRI model \cite{vetterli2002sampling} to multidimensional signals.

The main focus of this paper is to extend the multidimensional FRI model \cite{pan2013sampling} to enable super-resolution MRI. While the multidimensional FRI model \cite{pan2013sampling} is very powerful, it has some limitations that restricts its direct applicability to MRI. First of all, the complex analytic signal model introduced in \cite{pan2013sampling} is too restrictive and is not applicable to most MR images. We generalize the signal model to piecewise polynomial and harmonic functions, which are better suited to represent practical signals. The annihilation conditions central to the scheme are not exactly satisfied in the presence of model mismatch and noise. A Cadzow iterative procedure was used in \cite{pan2013sampling} to denoise the data before estimating the curve coefficients. This approach is computationally demanding, and requires some knowledge of the underlying model order. We introduce a novel algorithm based on averaging vectors in the null space of the equations that is robust to noise and model-mismatch, and is computationally efficient. 

Our reconstruction algorithm proceeds in two steps: First, we estimate a super-resolved spatial mask whose zeros correspond to the edges in the image. Second, we discretize the mask at the desired resolution and use the discrete spatial weights in a weighted total variation. We compare the efficacy of this two step strategy against classical discrete total variation regularized recovery of numerical phantoms from their exact Fourier samples. 

\section{SUPER-RESOLUTION OF EDGE IMAGES}
We will show under certain conditions it is possible to super-resolve the edges of an image from its low-frequency Fourier samples. In general we cannot hope to recover the edge set unless we assume certain constraints on its geometry. As in \cite{pan2013sampling} we assume the edge sets to be the zero set of a bandlimited periodic trigonometric polynomial $\mu(\mbf r)$ on $[0,1]^2$,
\begin{equation}
	C:~~\underbrace{\sum_{\mbf k\in\Lambda}\, \widehat{\mu}[\mbf k]\, e^{j2\pi \langle \mbf k, \mbf r\rangle}}_{\mu(\mbf r)} = 0
	\label{eq:annihilation}
\end{equation}
where $\Lambda$ is any finite index set,
and $\widehat\mu[\mbf k]$ are any complex coefficients. We call such sets $C$ \emph{trigonometric curves}. As noted in \cite{pan2013sampling}, the set of trigonometric curves have a rich topology, and for large enough bandwidth, they can approximate arbitrary curves to any desired accuracy. 

As a first approximation, we consider images that are \emph{piecewise constant}, meaning the image can be expressed as finite linear combinations of functions of the form
\begin{equation*}
1_\Omega(\mathbf r) = 
\begin{cases}
1 & \text{if } \mbf r = (x,y) \in \Omega\\
0 & \text{else},
\end{cases}
\end{equation*}
where $\Omega$ is a simple region in $[0,1]^2$ with piecewise smooth boundary $\partial \Omega$ given as $\{\mu = 0\}$ for some $\mu$ as in \eqref{eq:annihilation}. The Fourier transform of $1_\Omega$ is given as
\begin{align}
\nonumber \widehat 1_\Omega(\bs \omega) & =  \int_{\Omega} e^{-j \langle \bs \omega, \mbf r\rangle} d\mbf r = -\frac{1}{j \omega_x} \int_\Omega \partial_x \left[e^{-j \langle \bs \omega, \mbf r\rangle}\right] d\mbf r\\
 \label{eq:ftx}
 & = -\frac{1}{j \omega_x}\oint_{\partial\Omega}e^{-j \langle \bs \omega, \mbf r\rangle}dy, \text{ or },\\
 \label{eq:fty}
 & = \frac{1}{j\omega_y}\oint_{\partial\Omega}e^{-j\langle \bs \omega, \mbf r\rangle}dx,
\end{align}
where the last two equations follow by Green's theorem. Under the Fourier domain relations
\[
 \partial_x 1_\Omega (\mbf r) {\leftrightarrow} -j\omega_x \widehat{1_\Omega}(\bs \omega),\quad \partial_y 1_\Omega (\mbf r) \leftrightarrow -j\omega_y \widehat{1_\Omega}(\bs \omega)
\] the partial derivatives of $1_\Omega$ can be interpreted (in a distributional sense) as a continuous stream of Diracs in the spatial domain supported on $\partial\Omega$. Thus, formally, we should have $\mu \cdot \partial_x 1_\Omega = \mu \cdot \partial_y 1_\Omega= 0$, and so we say $\mu$ acts as an \emph{annihilating mask} for the partial derivatives of $1_\Omega$. This can be established rigorously in the Fourier domain to give the following:
\begin{proposition}
Let $1_\Omega$ be as above with boundary $\partial\Omega$ given as a trigonometric curve $C: \{\mu(\mbf r) = 0\}$. Then the Fourier transforms of the partial derivatives of $1_\Omega$ are annihilated by convolution with $\widehat{\mu}[\mbf k]$, that is
\begin{equation}
	\sum_{\mbf k \in \Lambda} \widehat{\mu}[\mbf k]\, \widehat{f}(\bs\omega - 2\pi\mbf k) = 0,\quad \text{for all } \bs \omega \in \mathbb{R}^2,
	\label{eq:prop1}	
\end{equation}
where $\widehat{f}(\bs \omega) =  - j \omega_x \widehat{1_\Omega}(\bs \omega)$ or $\widehat{f}(\bs \omega) := - j \omega_y \widehat{1_\Omega}(\bs \omega)$.
\end{proposition}

The above proposition shows that in principle it is possible to recover the edge set $C : \{\mu = 0\}$ of $1_\Omega$ by solving the linear system of equations \eqref{eq:prop1} for $\widehat{\mu}$. This is provided we have a sufficient number of low-pass samples to make the system \eqref{eq:prop1} determined. We investigate methods for solving for the filter coefficients $\widehat{\mu}$ in the following section.

We now consider the case where the image is assumed to be \emph{piecewise linear}, meaning it can be written as a linear combination of functions of the form
\begin{equation}
g(\mathbf r) = \begin{cases}
L(\mbf r) &  \text{if } \mbf r=(x,y) \in \Omega\\
0 & \text{else}
\end{cases}
\end{equation}
where $L$ is any affine function $L(\mbf r) = \mbf a^T \mbf r + b$, for $\mbf a \in \mathbb{C}^2$, $b \in \mathbb{C}$. Intuitively, any second derivative of $g$ should vanish except on $\partial\Omega : \{\mu(\mbf r) = 0\}$, where it will act like the derivative of a Dirac. Accordingly, we can show $\nu = \mu^2$ is an annihilating mask for any second derivative of $g$, since both $\nu$ and $\nabla \nu = 2\mu \nabla \mu$ vanish on $\partial\Omega$. The following proposition expresses this fact in the Fourier domain:
 \begin{proposition}
  \label{prop:2}
 Let $g = L \cdot 1_\Omega$ where $\partial^2 L = 0$, for some second order differential operator $\partial^2$. Then the Fourier transform of $h = \partial^2 g$ is annihilated by convolution with $\widehat{\mu^2}[\mbf k]$ on its support $\Lambda'$, that is
 \[
		\sum_{\mbf k \in \Lambda'} \widehat{\mu^2}[\mbf k]\, \widehat{h}(\bs\omega - 2\pi\mbf k) = 0,\quad \text{for all } \bs \omega \in \mathbb{R}^2
 \]
\end{proposition}
Note that a similar claim holds with the Laplacian $\Delta$ in place of $\partial^2$ and when $L$ is any harmonic function, $\Delta L = 0$. Therefore, images that are piecewise harmonic will satisfy the same annihilation property in Prop.\ \ref{prop:2}. 

\section{ALGORITHMS}
\subsection{Computation of an annihilating filter}
\label{sec:alg}
Suppose we wish to find filter coefficients $\mbf c \in \mathbb{C}^{M}$ supported in $\Lambda$ that annihilate the low-frequency Fourier coefficients of the image derivatives $\widehat{\mbf d_j} \in \mathbb{C}^N$, $j=1,...,n$, that is
\[
	(\widehat{\mbf d_j}\ast \mbf c)[\mbf k] = 0\quad \text{ for all } \mbf k \in \Lambda,\quad j=1,...,n.
\]
This is equivalent to finding a non-zero vector $\mbf c$ in the nullspace of the each block Toeplitz matrix $\mbf T_i \in \mathbb{C}^{M\times M}$ corresponding to 2-D convolution with $\widehat{\mbf d_i}$. Setting $\mbf T = [\mbf T_1^T,...,\mbf T_n^T]^T$, one approach would be to solve the least squares problem 
\begin{equation}
	\min_{\mbf c} \|\mbf T \mbf c\|_2^2,\quad \text{subject to}\quad \mbf c[\bs 0] = 1. 
\label{eq:LS}
\end{equation}
While computationally efficient, this method is highly sensitive to noise. Moreover, when the model order $M$ is larger than strictly necessary, this can lead to spurious zeros in the reconstructed mask $\mu = \mathcal{F}^{-1}[\mbf c]$. 

In the case of noisy measurements, the authors of \cite{pan2013sampling} suggest first performing a Cadzow denoising \cite{cadzow1988signal} 
of the Toeplitz matrix $\mbf T$, then solving \eqref{eq:LS}. However, the computational demands of Cadzow's algorithm are significant in this case, since each iteration of the algorithm requires finding the SVD of a large convolution matrix. Moreover, we found Cadzow's algorithm was unable to correct for spurious zeros unless the model order reduction was chosen correctly, which is difficult to know in advance.

To reduce the impact of noise and avoid spurious zeros in our annihilating mask, we propose a procedure that averages a collection of elements in the (approximate) nullspace of $\mbf T$. First, consider the noiseless overdetermined case where $\mbf T$ is low-rank. To obtain a basis for the nullspace of $\mbf T$ we compute its SVD, $\mbf T = \mbf U \bs \Sigma \mbf V^H$, and collect the columns of $\mbf V$ corresponding to zero singular values, labeling them $\{\mbf v_i\}_{i=1}^P$. From each $\mbf v_i$ we can construct an annihilating mask $\mu_i = \mathcal{F}^{-1}[\mbf v_i]$. Individually each $\mbf \mu_i$ may contain extra zeros outside $C$, but the intersection of their zero sets is likely only to contain $C$, due to orthogonality constraints. Hence we may compute an improved annihilating mask $\bar{\mu}$ by forming the sum-of-squares average
\begin{equation}
	\bar{\mu}(\mbf r) = \frac{1}{P}\left(\sum_{i=1}^P|\mu_i(\mbf r)|^2\right)^{1/2}.
\label{eq:sumofsquares}
\end{equation}
In the presence of noise $\mbf T$ will only be approximately low-rank, and in general we may not know the minimal model order representing the edge image, meaning we cannot precisely identify the nullspace vectors $\{\mbf v_i\}$. Therefore we propose collecting the $\{\mbf v_i\}$ corresponding to singular values $\sigma_i$ less than some threshold $\tau = \delta\cdot\sigma_1$, where $\delta\in(0,1]$ is a constant depending on the noise level. In this case, forming the averaged mask in \eqref{eq:sumofsquares} has the additional benefit of smoothing out variations in each mask $\mu_i$ due to noise.

\begin{figure}[h!]
        \centering
        \subfloat[Least-Squares]{\includegraphics[width=0.145\textwidth]{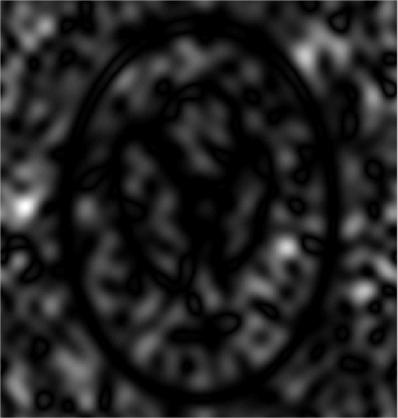}}~        
        \subfloat[Cadzow]{\includegraphics[width=0.145\textwidth]{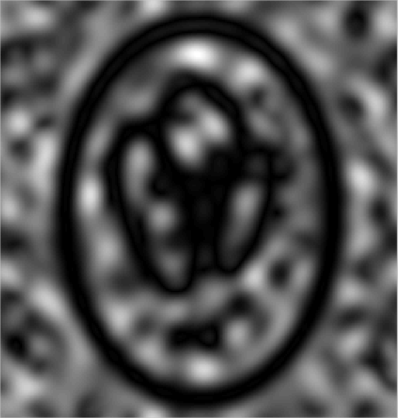}}~         
        \subfloat[Proposed]{\includegraphics[width=0.145\textwidth]{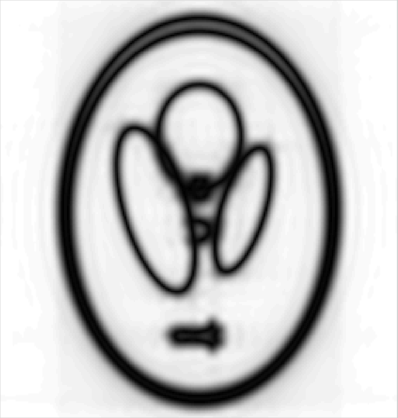}}
        \includegraphics[height=0.16\textwidth,trim = 100mm 10mm 0mm 0mm, clip]{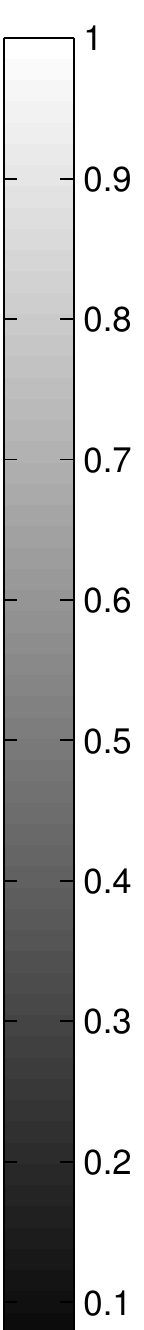}

        \subfloat[Least-Squares]{\includegraphics[width=0.145\textwidth]{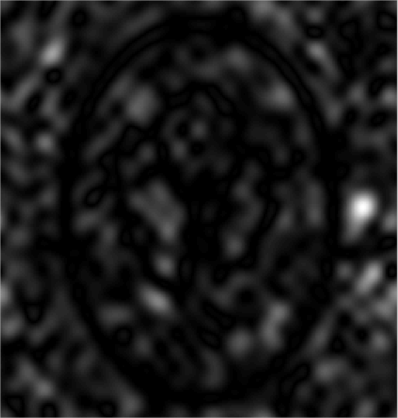}}~        
        \subfloat[Cadzow]{\includegraphics[width=0.145\textwidth]{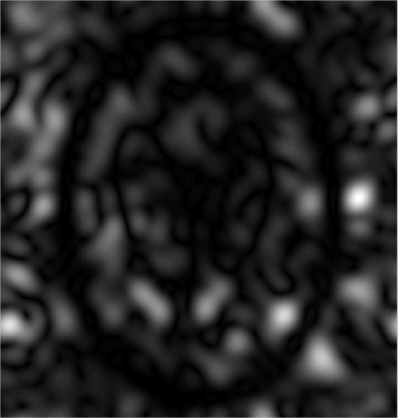}}~         
        \subfloat[Proposed]{\includegraphics[width=0.145\textwidth]{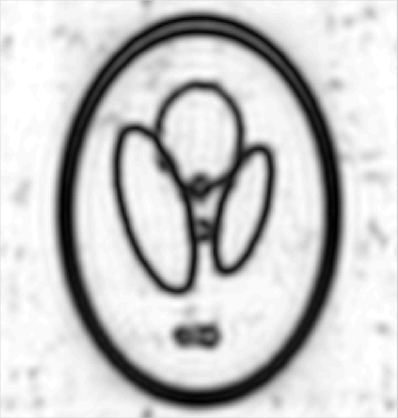}}
        \includegraphics[height=0.16\textwidth,trim = 100mm 10mm 0mm 0mm, clip]{images/colorbar_bw_trim.pdf}

        \caption{Edge annihilation masks for Shepp-Logan phantom using $61^2$ k-space samples. Top row shows a low-noise setting (SNR=25 dB), bottom row shows a high-noise setting (SNR=20 dB). The proposed method yields a significantly improved mask in both settings.} \label{fig:masks}
\end{figure}

In Fig.\ \ref{fig:masks} we compare the proposed method for computing an edge mask versus the least-squares and Cadzow denoising methods proposed in \cite{pan2013sampling} on the Shepp-Logan phantom in both low noise and high noise settings. Here we corrupt the k-space samples with mean-zero additive white Gaussian noise to have the indicated signal-to-noise ratio (SNR). We observe that the proposed method gives significantly improved masks in both cases.

\begin{figure*}[ht!]
\centering
\subfloat[Fully sampled, zoom]{
\begin{tikzpicture}[      
        every node/.style={anchor=south west,inner sep=0pt},
        x=1mm, y=1mm,
      ]   
     \node (fig1) at (0,0)
       {\includegraphics[width=0.18\textwidth,trim = 20mm 0mm 40mm 60mm, clip]{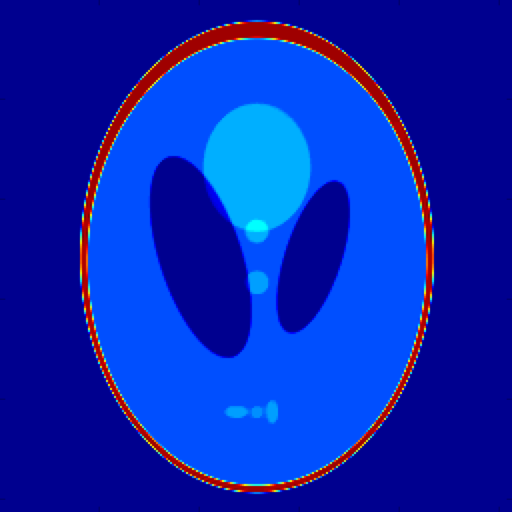}};
     \node [draw=white, thick] (fig2) at (0,0)
       {\includegraphics[width=0.06\textwidth]{images/SL_nonoise_orig_jet.png}};  
\end{tikzpicture}
}~~
\subfloat[TV, SNR=16.6dB]{
\begin{tikzpicture}[      
        every node/.style={anchor=south west,inner sep=0pt},
        x=1mm, y=1mm,
      ]   
     \node (fig1) at (0,0)
       {\includegraphics[width=0.18\textwidth,trim = 20mm 0mm 40mm 60mm, clip]{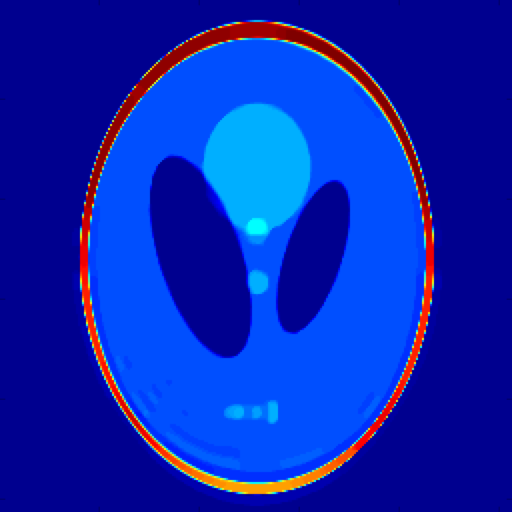}};
     \node [draw=white, thick] (fig2) at (0,0)
       {\includegraphics[width=0.06\textwidth]{images/SL_nonoise_TV_optlam_jet.png}};  
\end{tikzpicture}
}\hspace{-2mm}
\subfloat[Proposed, SNR=21.3dB]{
\begin{tikzpicture}[      
        every node/.style={anchor=south west,inner sep=0pt},
        x=1mm, y=1mm,
      ]   
     \node (fig1) at (0,0)
       {\includegraphics[width=0.18\textwidth,trim = 20mm 0mm 40mm 60mm, clip]{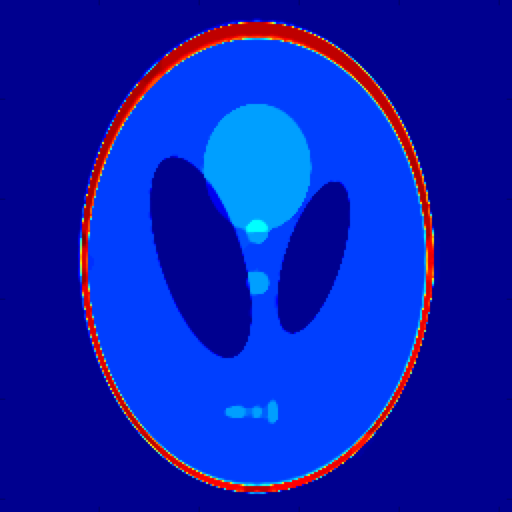}};
     \node [draw=white, thick] (fig2) at (0,0)
       {\includegraphics[width=0.06\textwidth]{images/SL_nonoise_WTV_jet.png}};  
\end{tikzpicture}
}~~
\subfloat[TV, SNR=14.6dB]{
\begin{tikzpicture}[      
        every node/.style={anchor=south west,inner sep=0pt},
        x=1mm, y=1mm,
      ]   
     \node (fig1) at (0,0)
       {\includegraphics[width=0.18\textwidth,trim = 20mm 0mm 40mm 60mm, clip]{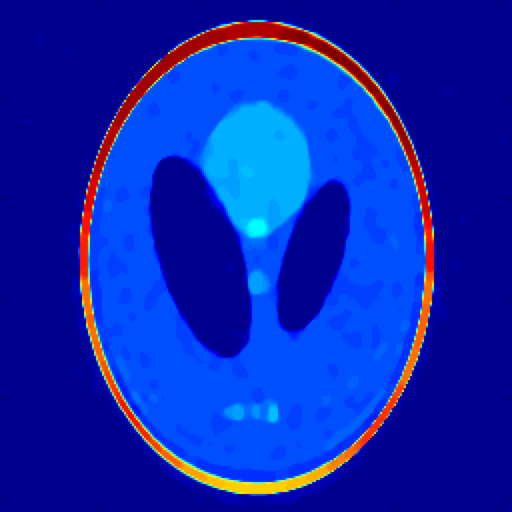}};
     \node [draw=white, thick] (fig2) at (0,0)
       {\includegraphics[width=0.06\textwidth]{images/SL_noisy_TV_optlam_jet.png}};  
\end{tikzpicture}
}\hspace{-2mm}
\subfloat[Proposed, SNR=20.1dB]{
\begin{tikzpicture}[      
        every node/.style={anchor=south west,inner sep=0pt},
        x=1mm, y=1mm,
      ]   
     \node (fig1) at (0,0)
       {\includegraphics[width=0.18\textwidth,trim = 20mm 0mm 40mm 60mm, clip]{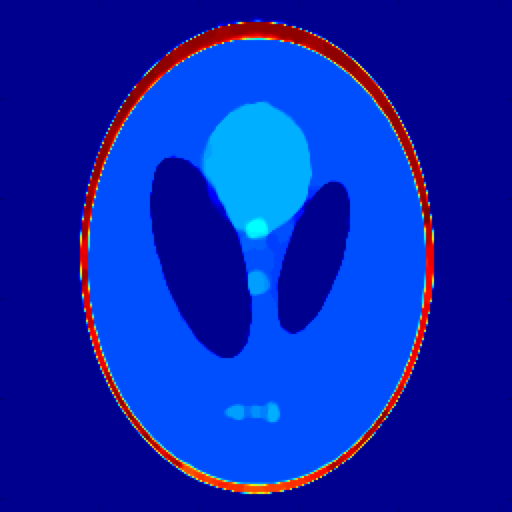}};
     \node [draw=white, thick] (fig2) at (0,0)
       {\includegraphics[width=0.06\textwidth]{images/SL_noisy_WTV_jet.png}};  
\end{tikzpicture}
}
\includegraphics[height=32mm,trim = 115mm 3mm 0mm 3mm, clip]{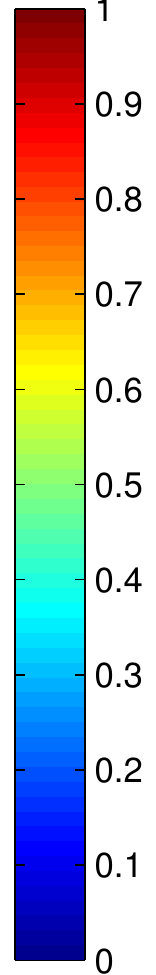}\\
\subfloat[Fully sampled, zoom]{
\begin{tikzpicture}[      
        every node/.style={anchor=south west,inner sep=0pt},
        x=1mm, y=1mm,
      ]   
     \node (fig1) at (0,0)
       {\includegraphics[width=0.18\textwidth,trim = 50mm 10mm 40mm 80mm, clip]{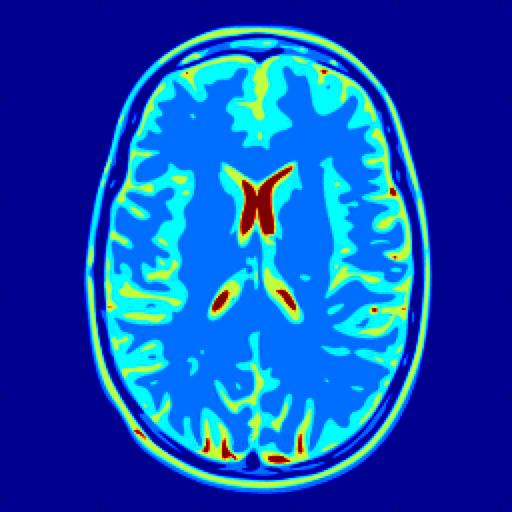}};
     \node [draw=white, thick] (fig2) at (0,0)
       {\includegraphics[width=0.06\textwidth]{images/brain_nonoise_smallfilter_orig_jet.png}};  
\end{tikzpicture}
}~~
\subfloat[TV, SNR=19.1dB]{
\begin{tikzpicture}[      
        every node/.style={anchor=south west,inner sep=0pt},
        x=1mm, y=1mm,
      ]   
     \node (fig1) at (0,0)
       {\includegraphics[width=0.18\textwidth,trim = 50mm 10mm 40mm 80mm, clip]{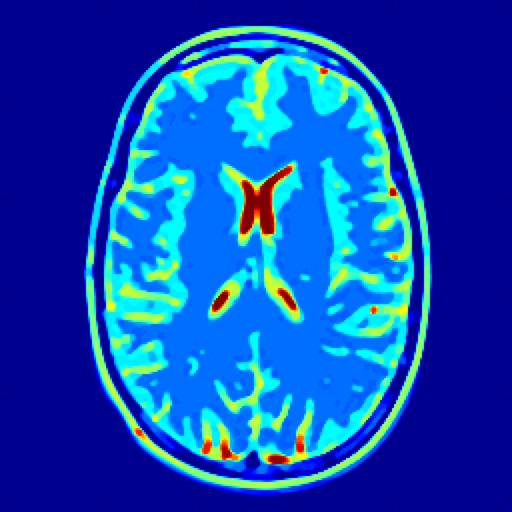}};
     \node [draw=white, thick] (fig2) at (0,0)
       {\includegraphics[width=0.06\textwidth]{images/brain_nonoise_smallfilter_TV_optlam_jet.png}};  
\end{tikzpicture}
}\hspace{-2mm}
\subfloat[Proposed, SNR=19.0dB]{
\begin{tikzpicture}[      
        every node/.style={anchor=south west,inner sep=0pt},
        x=1mm, y=1mm,
      ]   
     \node (fig1) at (0,0)
       {\includegraphics[width=0.18\textwidth,trim = 50mm 10mm 40mm 80mm, clip]{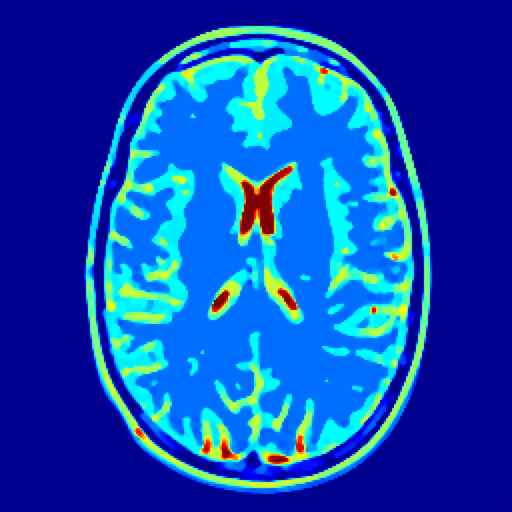}};
     \node [draw=white, thick] (fig2) at (0,0)
       {\includegraphics[width=0.06\textwidth]{images/brain_nonoise_smallfilter_WTV_jet.png}};  
\end{tikzpicture}
}~~
\subfloat[TV, SNR=16.9dB]{
\begin{tikzpicture}[      
        every node/.style={anchor=south west,inner sep=0pt},
        x=1mm, y=1mm,
      ]   
     \node (fig1) at (0,0)
       {\includegraphics[width=0.18\textwidth,trim = 50mm 10mm 40mm 80mm, clip]{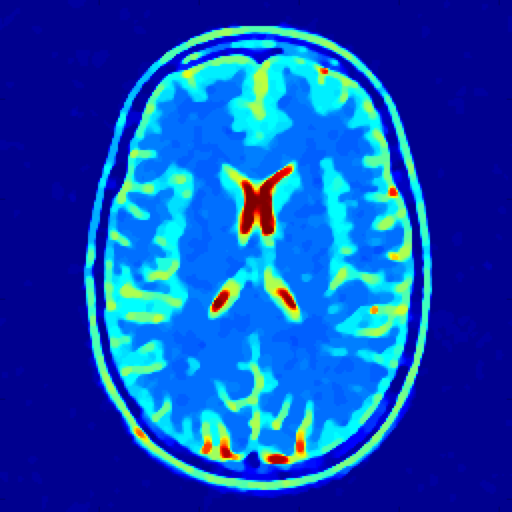}};
     \node [draw=white, thick] (fig2) at (0,0)
       {\includegraphics[width=0.06\textwidth]{images/brain_noisy_new_TV_optlam_jet.png}};  
\end{tikzpicture}
}\hspace{-2mm}
\subfloat[Proposed, SNR=17.1dB]{
\begin{tikzpicture}[      
        every node/.style={anchor=south west,inner sep=0pt},
        x=1mm, y=1mm,
      ]   
     \node (fig1) at (0,0)
       {\includegraphics[width=0.18\textwidth,trim = 50mm 10mm 40mm 80mm, clip]{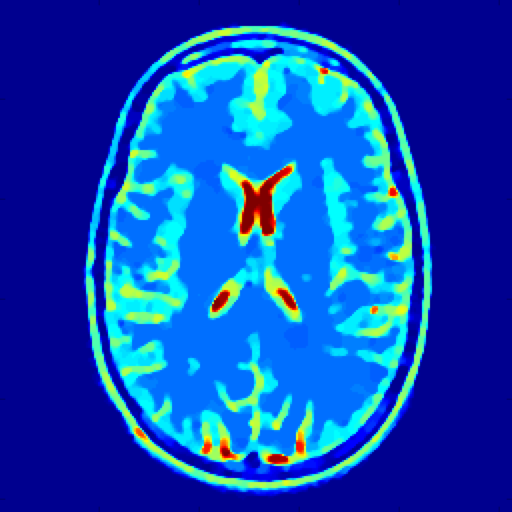}};
     \node [draw=white, thick] (fig2) at (0,0)
       {\includegraphics[width=0.06\textwidth]{images/brain_noisy_new_WTV_jet.png}};  
\end{tikzpicture}
}
\includegraphics[height=32mm,trim = 115mm 3mm 0mm 3mm, clip]{images/colorbar_jet_trim_me.pdf}
\caption{\small (Top row) Shepp-Logan phantom reconstructed onto a $256$$\times$$256$ spatial grid from $65\times 49$ center k-space samples ($\approx$ 20-fold undersampling) with no added noise in (b)\&(c) and 25dB added noise (d)\&(e). (Bottom row) Brain phantom reconstructed onto $256$$\times$$256$ spatial grid reconstructed from $97\times 97$ center k-space samples ($\approx$ 7-fold undersampling) with no added noise in (g)\&(h) and 30dB added noise (i)\&(j). Note the proposed scheme gives sharper reconstructions with fewer aliasing artifacts over standard TV. }
\label{fig:SL_K10}
\end{figure*}

\subsection{Reconstruction algorithm}
Following \cite{pan2013sampling}, we propose a two-stage algorithm for recovering MR images. The first stage is to estimate an annihilating mask $\mu$ using the procedure in the previous subsection. Note that this first step can be carried out in a resolution-independent manner---the mask $\mu$ records the edge locations with infinite resolution. In the second stage, we discretize the mask $\mu$ to the desired resolution as a set of spatial weights $\mbf W_\mu$ and perform a weighted total variation (TV) recovery:
\begin{equation}
	\min_{\mbf x} \|\mbf A \mbf x - \mbf b\|_2^2 + \lambda\,\|\mbf W_\mu \cdot |\nabla \mbf x|\|_1
\label{eq:tvopt}
\end{equation}
where $\mbf x$ the discrete image to be recovered, $\mbf A$ is the Fourier undersampling operator, $\mbf b$ is the vector of noisy low-pass Fourier measurements, $\lambda$ is a regularization parameter, and $|\nabla \mbf x|$ denotes the magnitude of the discrete gradient of $\mbf x$. In the case where a higher order signal model is assumed, we could also replace the weighted TV penalty with a weighted higher degree TV (HDTV) penalty \cite{hu2012higher}.

\section{RESULTS}
In this section we demonstrate the utility of our proposed recovery scheme for single-coil MR image recovery from low-pass k-space samples. For simplicity, we focus on the first-order case, using piecewise constant phantoms. In order to obtain resolution-independent simulations, we calculate k-space data directly using the analytical MRI phantoms derived in \cite{guerquin2012realistic}. We simulate a noisy acquisition by adding complex white Gaussian noise with standard deviation $\sigma$ to the computed k-space samples.

We collect Cartesian center k-space samples located within a rectangular window $[-2K,2K]$$\times$$[-2L,2L]$, and solve for an edge mask using the procedure outlined in Sec.\ \ref{sec:alg}. We then recover the image using the weighted TV scheme \eqref{eq:tvopt}, and compare against standard TV, i.e.\ \eqref{eq:tvopt} with $W_\mu=\bs 1$. In both cases we tune the regularization parameter $\lambda$ to optimize the signal-to-noise ratio (SNR), defined as $\text{SNR} = 20\log_{10}(\|\mbf x_0\|_2/\|\mbf x - \mbf x_0\|_2)$ where $\mbf x$ is the reconstructed image, and the ground truth $\mbf x_0$ is a TV reconstruction of the image from fully-sampled k-space.

Figure \ref{fig:SL_K10} shows the reconstructions obtained at a resolution of $256$$\times$$256$ from $65\times 49 = 3185$ k-space samples in the case of the Shepp-Logan phantom, and from $97\times 97 = 9409$ noisy k-space samples for the brain phantom, in both noiseless and noisy setting. We observe that in the standard TV reconstructions, using a low regularization weight recovers much of the fine detail but results in significant aliasing artifacts. However, increasing the regularization weight enough to eliminate the aliasing results in oversmoothing. In contrast, performing a weighted TV with an edge mask allows us to simultaneously recover the fine details while also suppressing aliasing artifacts.

\section{CONCLUSIONS}
We extend the theory in \cite{pan2013sampling} to higher-order signal models, and propose a more robust and efficient means of computing super-resolved edge images. Our experiments on analytical phantoms demonstrate the potential of the proposed scheme for super-resolution MRI. 







\bibliographystyle{IEEEbib}
\bibliography{IEEEabrv,root}

\end{document}